\documentclass[final]{cvpr}

\usepackage{times}
\usepackage{epsfig}
\usepackage{graphicx}
\usepackage{amsmath}
\usepackage{amssymb}
\usepackage{psfrag}
\usepackage[linesnumbered,ruled]{algorithm2e}

\usepackage[pagebackref=true,breaklinks=true,colorlinks,bookmarks=false]{hyperref}

\def\confYear{CVPR 2021}

\begin{document}

\title{Efficient Sampling for Predictor-Based Neural Architecture Search}

\author{Lukas Mauch, Stephen Tiedemann, Javier Alonso Garcia, Bac Nguyen Cong, Kazuki Yoshiyama,\\
Fabien Cardinaux, Thomas Kemp\\
Sony Europe B.V., Stuttgart, Laboratory 1\\
{\tt\small \{Lukas.Mauch, Stephen.Tiedemann, Javier.AlonsoGarcia, Bac.NguyenCong, Kazuki.Yoshiyama}\\ 
{\tt\small Fabien.Cardinaux, Thomas.Kemp\}@sony.com}}

\maketitle

\begin{abstract}
Recently, predictor-based algorithms emerged as a promising approach for neural architecture search (NAS).
For NAS, we typically have to calculate the validation accuracy of a large number of Deep Neural Networks (DNNs), what is computationally complex. Predictor-based NAS algorithms address this problem. They train a proxy model that can infer the validation accuracy of DNNs 
directly from their network structure. During optimization, 
the proxy can be used to narrow down the number of architectures for which the true validation
accuracy must be computed, what makes predictor-based algorithms sample efficient. 
Usually, we compute the proxy for all DNNs in the network search 
space and pick those that maximize the proxy as candidates for optimization. 
However, that is intractable in practice, because the search spaces are often very large and contain billions of network architectures. The contributions of this paper are threefold:
1) We define a sample efficiency gain to compare different predictor-based NAS algorithms. 
2) We conduct experiments on the NASBench-101 dataset and show that the sample efficiency of predictor-based algorithms decreases dramatically if the proxy is only computed for a subset of the search space. 
3) We show that if we choose the subset of the search space on which the proxy is evaluated in a smart way, the
sample efficiency of the original predictor-based algorithm that has access to the full search space can be regained. 
This is an important step to make predictor-based NAS algorithms useful, in practice.
\end{abstract}

\section{Introduction}
\label{sec:introduction}
In the recent decade, the introduction of Deep Neural Networks (DNNs) 
revolutionized machine learning and caused a shift of paradigm, away from
feature and kernel engineering, towards network architecture design. 
While many DNNs that perform well across various machine learning tasks have 
emerged from a manual design, constructing DNNs that are highly accurate and resource efficient at the same time
has proven to be a tedious task that requires much intuition, experience and time. 
Fortunately, recent works on Neural Architecture Search (NAS) suggest that 
we can discover new network architectures in an automated way.
In fact, as discussed in \cite{zoph2016nas, liu2018darts, cai2018proxylessnas, cai2019ofa, white2019bananas}, 
NAS has been applied to many deep learning tasks such as image recognition and language modelling with great success.

NAS algorithms typically operate on a pre-defined
search space $\mathcal{S}$ that contains up to billions of candidate networks. 
Architecture search means to pick the network architecture with the smallest validation error.
Mathematically, this corresponds to the following bi-level optimization problem
\begin{align}
    \mathcal{A}^* & = \arg \min_{\mathcal{A}} J_v(\mathcal{A}, \mathbf{w}^*) \\
    \text{s.t. } & \mathbf{w}^* = \arg \min_{\mathbf{w}} J_t(\mathcal{A}, \mathbf{w}),
\end{align}
where $\mathcal{A}$ defines the network architecture, 
$\mathbf{w}$ are the corresponding network weights and $J_t(\cdot)$ and
$J_v(\cdot)$ are the training loss and the validation error, respectively.

Solving this optimization problem is challenging in practice. 
To optimize $\mathcal{A}$, we need to solve a combinatorial optimization problem for which exhaustive search
is not tractable. In particular, the calculation of the validation error $J_v(\mathcal{A}, \mathbf{w}^*)$ involves
the computation of the optimal network weights $\mathbf{w}^*$, 
meaning that each candidate network must be trained until convergence. 

In general, this complexity issue can be addressed in two ways:
1) Speed up the inner optimization problem, i.e., reduce the time it takes to evaluate validation cost $J_v(\mathcal{A}, \mathbf{w}^*)$ for a given architecture $\mathcal{A}$. 
2) Use sample efficient algorithms to optimize the outer problem, 
such that $J_v(\mathcal{A}, \mathbf{w}^*)$ must only be evaluated for as few candidate architectures as possible.

One-shot, progressive shrinking or single-stage NAS 
algorithms, as proposed by \cite{cai2019ofa} and \cite{yu2020bignas}, follow the first approach. 
Rather than training single candidate networks in the search 
space until convergence, what requires much time, they propose to
train one big super-net (or single-stage-model) that contains all networks in $\mathcal{S}$ as sub-networks. 
Once the super-net is trained, $\mathbf{w}^*$ can be calculated for any $\mathcal{A}$ by dropping all parts that do not
belong to $\mathcal{A}$. Training one super-net rather than thousands of stand alone networks is desirable in practice, because it can be done much faster.
However, as discussed in \cite{yu2020bignas} defining and training such a super-net proves to be a challenging task on its own
that requires many tricks. Further, there exist controversial results 
whether networks picked from a super-net preserve the 
same accuracy ranking as networks that have been trained individually.

\begin{figure}
    \centering
    \includegraphics[width=\linewidth]{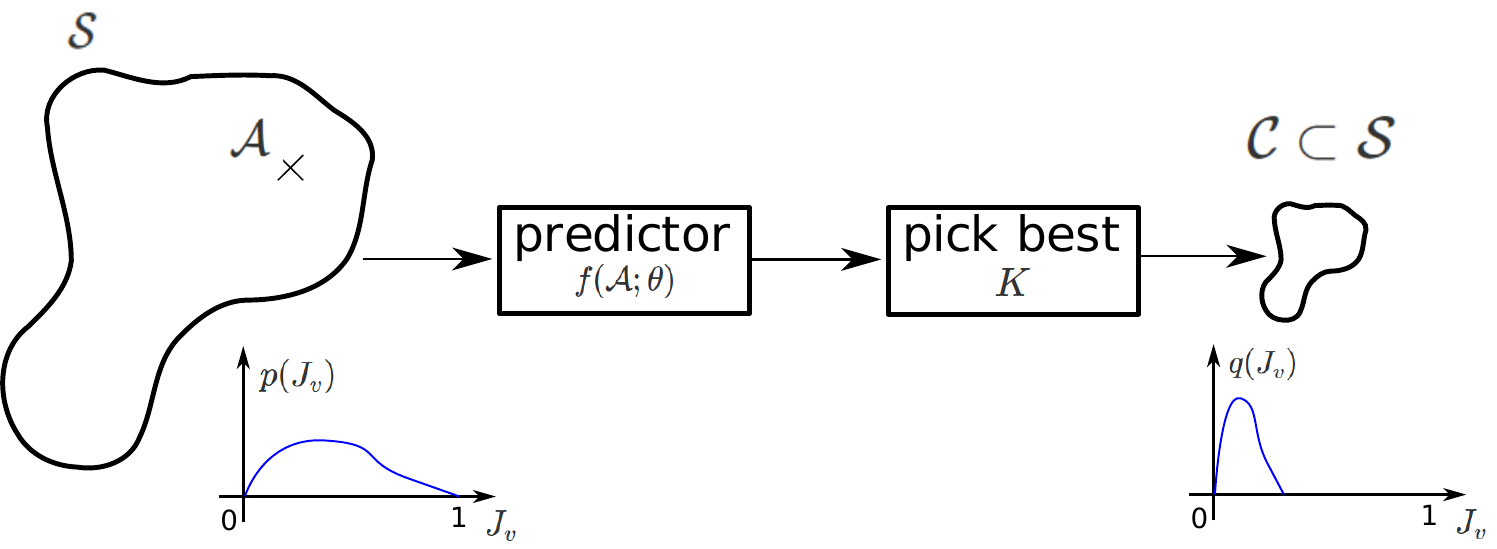}
    \caption{Predictor-based NAS algorithms use a proxy $f(\mathcal{A};\theta)$ of the true validation error to narrow down the search space $\mathcal{S}$, i.e., they propose a set of candidate architectures $\mathcal{C} \subset \mathcal{S}$ by discarding architectures with a large predicted validation error.}
    \label{fig:pb_nas}
\end{figure}

As discussed in \cite{wen2019neural, ning2020generic}, predictor-based NAS algorithms follow the second approach.
They use a predictor to increase the sample efficiency of the NAS algorithm. 
As shown in Figure~\ref{fig:pb_nas}, let $f(\mathcal{A}; \mathbf{\theta})$
be a predictor with parameters $\mathbf{\theta}$, that 
is trained as a proxy for the true validation error $J_v(\mathcal{A}, \mathbf{w}^*)$ 
of a network architecture $\mathcal{A}$, using either a regression or a ranking loss.
During optimization, $f(\mathcal{A}; \mathbf{\theta})$ is used to propose a set of candidate architectures 
$\mathcal{C} \subset \mathcal{S}$ by discarding architectures with a large predicted validation error. 
Let $p(J_v)$ and $q(J_v)$ be the distribution of the true validation error of network architectures from $\mathcal{S}$ and
$\mathcal{C}$, respectively. The main idea of predictor-based NAS is, that if $f(\mathcal{A}; \mathbf{\theta})$ is trained
well, $q(J_v)$ is a much narrower distribution with considerably more probability mass close to $J_v = 0$. 
Hence, compared to the full search space $\mathcal{S}$, we will have to 
train and evaluate considerable less networks from the candidate set $\mathcal{C}$,
until we observe reasonable low validation errors $J_v(\mathcal{A}, \mathbf{w}^*)$.
Usually, the parameters of $f(\mathcal{A}; \mathbf{\theta})$ are trained once, 
using a small subset of $\mathcal{S}$ for which the true validation error is known. 
This is for example done in \cite{wen2019neural}. However, the parameters of $f(\mathcal{A}; \mathbf{\theta})$ and the
candidate set $\mathcal{C}$ can also be refined each time when new validation errors are observed.

A major problem that makes predictor-based NAS algorithms hard to apply in practice is that they do not scale well to large search spaces. In practice, the search space size
grows exponentially with the number of layers and the number of possible operations that we allow.
Consider for example that we would like to search for the optimal number of feature maps in a ResNet50, allowing $64,...,512$ feature maps in each layer. In this case, the search space would consist of $448^{50}$ different network architectures. Therefore, evaluating $f(\mathcal{A};\theta)$ on the whole search space $S$ to construct the candidate set
$\mathcal{C}$ is not an option. In this paper, we address this important issue and give a possible solution.

Our contributions are threefold: 
1) We define a sample efficiency gain to compare different predictor-based NAS algorithms. 
Depending on the distribution of the validation error, we calculate the expected number of network architectures $\tau(J_{target})$ that we need to train and evaluate until we observe a validation error that is better or equal than a desired target $J_{target}$. We do this for both, the original search space $\mathcal{S}$ and for the candidate set $\mathcal{C}$. The sample efficiency gain can then be calculated from the related change of $\tau(J_{target})$. 
2) In practice, we can calculate $f(\mathcal{A}; \theta)$ only for a small $\mathcal{S}' \subset \mathcal{S}$, in order to propose the set of candidates $\mathcal{C}$. We conduct experiments on the NASBench-101 dataset and show that the sample efficiency gain of predictor-based algorithms decreases dramatically if $|\mathcal{S}'| << |\mathcal{S}|$.
3) We show that if we choose $\mathcal{S}'$ in a smart way, the
sample efficiency of the original predictor-based algorithm that evaluates $f(\mathcal{A}; \theta)$ on the full search space $\mathcal{S}$ can be recovered. This is an important step to make predictor-based NAS algorithms useful, in practice.

The paper is structured as follows. 
In section~\ref{sec:related_work}, we review the most important related work. 
In section~\ref{sec:pb_nas}, we introduce the predictor-based NAS algorithm and define the 
graph convolutional network (GCN) that we use as the predictor for all our experiments.
In section~\ref{sec:gain}, we define the sample efficiency gain that we use to compare different predictor-based NAS algorithms.
Finally, in section~\ref{sec:experiments}, we conduct our experiments on the NASBench-101 dataset.

\section{Related Work}
\label{sec:related_work}
Many different approaches to improve the sample efficiency of NAS algorithms have been proposed.
Among them are reinforcement learning (RL) based \cite{zoph2016neural, guo2020breaking}, 
Bayesian optimization (BO) based \cite{white2019bananas}, 
gradient based \cite{liu2018darts, lian2019towards}, 
evolutionary \cite{liu2017hierarchical, real2019regularized} 
and predictor-based \cite{wen2019neural, ning2020generic} NAS algorithms.
Predictor-based NAS, are very similar to Bayesian optimization (BO), in the sense that
they rely on a proxy model of the actual cost function to determine highly probable candidate points \cite{brochu2010tutorial}.

The idea of predictor-based NAS has been studied in many papers 
\cite{wen2019neural, ning2020generic}. 
\cite{wen2019neural} proposed to train a graph convolutional network (GCN) as introduced by \cite{kipf2016semi} as a proxy for the true validation accuracy of a network architecture. In their experiments on the NASBench101 dataset \cite{ying2019-bench101}, they train the prediction model
with a regression loss on 175 architecture/accuracy pairs. They use the trained model
to predict the accuracy of all remaining architectures in the dataset and picking the networks with
the best predicted accuracies. This results in a $20 \times$ higher sample efficiency, 
compared to random architecture search. In average, they approach the architecture with the optimal validation accuracy after $1000$ samples. In experiments on ImageNet \cite{russakovsky2015imagenet}, they reach comparable performance to ProxylessNAS \cite{cai2018proxylessnas}. 

\cite{ning2020generic} compare the performance of different predictor models that are either trained with
regression or ranking losses. They argue that GCNs perform much better then previously proposed 
auto-regressive models that encode the network architecture in a sequential way. This is plausible, because
the output of GCNs is invariant if it gets isomorph network graphs as an input. Further, they propose an improved 
predictor model called GATES and show that predictor models perform best if they are trained with a ranking loss.

However, none of these papers addressed the inherent problem that predictor-based algorithms still do not 
scale well to huge search spaces. Although the proxy can be evaluated much faster
then the true validation error, it still requires a considerable effort to propose a good set of candidate architectures if
the search space is huge and contains billions of network architectures. 
We study this problem and point towards a possible solution that makes predictor-based NAS algorithms applicable to large search spaces.

\section{Predictor-based NAS}
\label{sec:pb_nas}

Algorithm~\ref{alg:pb_nas} summarizes the predictor-based NAS algorithm that we use in all our experiments. 
We start with an initial training set $\mathcal{T}_0$ of network architectures for which we have observed the true evaluation error. In each iteration $t=1,...,T$, the training set $\mathcal{T}_{t-1}$ is used to train a predictor $f(\mathcal{A}; \theta)$. That predictor is later used to draw
a set of $K$ candidate architectures from $\mathcal{S}$. We denote this candidate set with $\mathcal{C}$. In particular, we sort all architectures in $\mathcal{S}$ according to the related predictor output $\mathcal{F}$. Then, the best $K$ architectures are removed from $\mathcal{S}$ \footnote[1]{For comparisson, in algorithm~\ref{alg:pb_nas} pick\_best\_K not only returns the best $K$ architectures, but also removes them from $\mathcal{S}$.}. They are trained, evaluated and added to the training set $\mathcal{T}_{t+1}$,
that is used to train the predictor in the next iteration.

\begin{algorithm}
    \caption{The predictor-based NAS algorithm.}
    \label{alg:pb_nas}
    
     $\mathcal{T}_0 \gets \mathrm{random\_pick\_}K(\mathcal{S})$ \\
     $\mathcal{Y}_0 \gets \{ J_v(\mathcal{A}, \mathbf{w}^*) : \mathcal{A} \in \mathcal{T}_0 \}$\\
     $y^*_0 \gets \min \mathcal{Y}_0$\\
    
    \While{$t=1; \:\: t \leq T$}{
         $\mathbf{\theta}^* \gets \arg \min_{\mathbf{\theta}} J_d(\mathcal{S}_{t-1}, \mathcal{Y}_{t-1}; \mathbf{\theta})$ \\
        
         $\mathcal{F} \gets \{ f(\mathcal{A};\mathbf{\theta}^*) : \mathcal{A} \in \mathcal{S} \}$ \\
        
         $\mathcal{C} \gets \mathrm{pick\_best\_}K(\mathcal{S}, \mathcal{F})$\\
        
         $\mathcal{T}_t \gets \mathcal{T}_{t-1} \cup \mathcal{C}$\\
        
         $\mathcal{Y}_t \gets \mathcal{Y}_{t-1} \cup \{J_v(\mathcal{A}, \mathbf{w}^*): \mathcal{A} \in \mathcal{C}\}$\\
        
         $y^*_t \gets \min \mathcal{Y}_t$
    }
\end{algorithm}

\subsection{GCNs for predictor-based NAS}
\label{sec:predictor}
As discussed in \cite{wen2019neural}, it is natural to represent DNN architectures as graphs and therefore to use a GCN as
the predictor. Although we use a slightly simpler setup than \cite{wen2019neural}, experiments on NASBench-101 show, that it has comparable performance and therefore can be regarded as a state-of-the-art predictor.

Let $\mathcal{A} = (\mathbf{A}, \mathbf{X})$ be a network architecture with $L$ layers, that is defined by the layer adjacency matrix $\mathbf{A} \in \mathbb{R}^{L \times L}$ and by the layer feature matrix $\mathbf{X} \in \mathbb{R}^{L \times d}$, where $[\mathbf{A}]_{ij} = 1$ if the output of layer $i$ is an input to layer $j$ and $[\mathbf{A}]_{ij} = 0$, otherwise. Further, the $i$-th row of $\mathbf{X}$ encodes properties of the $i$-th network layer. In our case, we only use one-hot encodings of the layer type. However, we can also think of richer representations that could also capture the width of the input and output tensors, the number of parameters or the number of operations that are needed to evaluate the layer.

As described in \cite{kipf2016semi}, the GCN maps the graph representation of an architecture to a real number, i.e., $f : \mathbb{R}^{L \times L} \times \mathbb{R}^{L \times d} \rightarrow \mathbb{R}$, which can be sorted in order to rank the architectures. Our predictor consists of $G-2$ graph convolutional layers that are followed by a pooling and a fully-connected layer. Similar to \cite{kipf2016semi}, each graph convolutional layer computes
\begin{align}
    \mathbf{X}_g = f_g(\mathbf{A}, \mathbf{X}_{g-1}; \mathbf{W}_l) = \Phi_g ( \tilde{\mathbf{D}}^{-\frac{1}{2}} \tilde{\mathbf{A}} \mathbf{X}_{g-1} \mathbf{W}_g ),
\end{align}
where $\mathbf{W}_g \in \mathbb{R}^{d_{g-1} \times d_g}$ are the layer weights, 
$\tilde{\mathbf{A}} = \mathbf{A} + \mathbf{I}$ is the adjacency matrix with added self-connections and
$\tilde{\mathbf{D}} = \mathrm{diag}(\tilde{\mathbf{A}} \mathbf{1})$ is the corresponding degree matrix
that is used to normalize the activations 
\footnote[2]{Note that \cite{kipf2016semi} proposed a slightly different parametrization 
$\Phi_g ( \tilde{\mathbf{D}}^{-\frac{1}{2}} \tilde{\mathbf{A}} \tilde{\mathbf{D}}^{-\frac{1}{2}} \mathbf{X}_{g-1} \mathbf{W}_g )$. We normalize $\tilde{\mathbf{A}}$
only by $\tilde{\mathbf{D}}^{-\frac{1}{2}}$. This way, the variance of $\mathbf{X}_{g}$ is neither amplified, nor attenuated over the layers. This has proven to be beneficial in practice.}. 
The output of the $G-2$ graph convolutional layers is pooled over the layer dimension to obtain 
a vector representation of $\mathcal{A}$, i.e.,
\begin{align}
    [\mathbf{x}_{G-1}]_i = \frac{1}{L}\sum_{l=1}^L [\mathbf{X}_l]_{l,i},
\end{align}
where $\mathbf{x}_{G-1} \in \mathbb{R}^d$ is the output vector. The output of the GCN is finally calculated by a fully-connected output layer with just one neuron, that maps $\mathbb{R}^d \rightarrow \mathbb{R}$. The whole network computes $ f(\mathbf{A}, \mathbf{X}; \mathbf{\theta}) = (f_G \circ f_{G-1} \circ \hdots \circ f_1) (\mathbf{A}, \mathbf{X}; \mathbf{\theta})$. 
Note, that the output of the GCN is invariant if the layers of $\mathcal{A}$ are permuted. In particular, if $\mathbf{P} \in \mathbb{R}^{L \times L}$ is a permutation matrix,
$f(\mathbf{A}, \mathbf{X}; \mathbf{\theta}) = f(\mathbf{P} \mathbf{A} \mathbf{P}^T, \mathbf{P} \mathbf{X}; \mathbf{\theta})$. This is desirable, because such a permutation of the layers results in exactly the same DNN architecture and therefore should not change the output of the predictor.

As given in algorithm~\ref{alg:pb_nas}, we train the GCN with a pairwise cross-entropy loss that is minimized over the training set $\mathcal{T}_t \subset \mathcal{S}$, in each iteration. In particular, for a pair of architectures $(\mathcal{A}, \mathcal{A}') \in \mathcal{T}_t \times \mathcal{T}_t$, we define the probability that $\mathcal{A}$ has a larger validation accuracy then $\mathcal{A}'$, i.e.,
\begin{align}
    p(y|\mathcal{A}, \mathcal{A}')\propto \exp\{y(f(\mathbf{A}, \mathbf{X}; \mathbf{\theta}) + (1-y)f(\mathbf{A}', \mathbf{X}'; \mathbf{\theta})\}, \label{eq:likelihood}
\end{align}
with the random variable
\begin{align}
   y = \begin{cases}
        1 , & J_v(\mathcal{A}, \mathbf{w}^*) \geq J_v(\mathcal{A}, {\mathbf{w}'}^*)\\ 
        0 , & \text{others}
    \end{cases}.
\end{align}
The pairwise cross-entropy loss therefore is defined as
\begin{align}
    J_{d}(\mathcal{S}', \mathcal{Y}'; \mathbf{\theta}) = - \mathrm{E}_{\mathcal{A}, \mathcal{A}'}[\ln p(y| \mathcal{A}, \mathcal{A}')], \:\: \mathcal{A} \neq \mathcal{A}',
\end{align}
where $\mathcal{Y}' = \{J_v(\mathcal{A}, \mathbf{w}^*) : \mathcal{A} \in \mathcal{T}_t \}$ 
contains the true observed validation accuracy for the architectures in $\mathcal{T}_t$. 
Note, while \cite{wen2019neural} trains the predictor only once at the beginning of the optimization, 
using a fixed training set, we continuously increase the size of 
$\mathcal{T}_t$ and update the parameters of $f(\mathcal{A}; \theta)$, when new network architectures
have been trained and evaluated.

\subsection{Predictor-based NAS and large search spaces}
\label{sec:pb_sampling}
In practice, the search space $\mathcal{S}$ can contain billions of different network architectures.
When calculating the candidate set $\mathcal{C}$, it is therefore often impossible to evaluate 
$\mathcal{F} = \{f(\mathcal{A}; \theta): \mathcal{A} \in \mathcal{S} \}$. A solution to this problem is
to exchange $\mathcal{S}$ with $\mathcal{S}' \subset{S}$, where$|\mathcal{S}| << ,|\mathcal{S}'|$.
There are many different ways how to sample $\mathcal{S}'$. In this paper, we investigate three of them,
namely uniform, maximum-likelihood (ML) based and evolutionary sampling.

\emph{Uniform sampling} is the most straight forward idea how we can reduce $\mathcal{S}$. 
Let $N'$ be the desired number of elements in $\mathcal{S}'$. 
We first enumerate all elements in $\mathcal{S}$ and then draw $N'$ elements from
$\mathcal{S}$ without replacement, using an equal sampling probability for each network architecture.
This approach has also been used in \cite{wen2019neural}. Clearly, this method will have a negative influence on the performance of the predictor-based algorithm. Even if $f(\mathcal{A}; \theta)$ is trained well, meaning that we can use it to correctly choose the best $K$ architectures from $\mathcal{S}'$, the performance of the
NAS algorithm is limited by the probability that $\mathcal{S}'$ contains any network architecture with a low validation error. Depending on $N'$, this can cause that the performance of the NAS algorithm reduces to the performance of
random search.

\emph{Maximum likelihood based sampling} is another obvious idea to sample $\mathcal{S}'$. 
Here, we optimize
\begin{align}
    \hat{\mathcal{A}} &= \arg \max_{\mathcal{A}} p(y=1|\mathcal{A}, \mathcal{A}')\\
                      &= \arg \max_{\mathcal{A}} f(\mathcal{A}; \theta^*),
\end{align}
using gradient descent from $N'$ different random initial points. Because $f(\mathcal{A};\theta)$ is a highly nonlinear function, gradient descent will end up in a different local minimum of $f(\mathcal{A};\theta)$ for each random initial point from which we start.  Optimizing $\mathcal{A}=(\mathbf{A}, \mathbf{X})$ means to optimize the adjacency matrix $\mathbf{A}$ and the vertice features $\mathbf{X}$.
Because $\mathbf{A}$ and $\mathbf{X}$ contain binar values and one-hot encodings, respectively, gradient descent cannot be applied right away. We use continuous valued shadow matrices for $\mathbf{A}$ and $\mathbf{X}$ that are mapped to their binar and one-hot encoded equivalents, by thresholding and maximum calculation. Then, we apply the straight through gradient estimator (STE) to propagate through all non-differentiable operations. 

Compared to uniform sampling, this method uses the shape of $f(\mathcal{A};\theta)$ to guide the sampling. In particular, it is very unlikely that $\mathcal{S}'$ will contain a bad network architecture, because each sample already optimizes  $f(\mathcal{A};\theta)$. However, ML based sampling requires us to perform many gradient update steps. Therefore, has a much larger computational complexity than uniform sampling.

\emph{Evolutionary sampling} of $\mathcal{S}'$ is the third and last method that we investigate in this paper.
Here, we want to exploit, that the validation error $J_v(\mathcal{A}, \mathbf{w}^*)$ is a smooth function over $\mathcal{A}$, meaning that network architectures with a similar network graph also have a similar validation error.
Evolutionary sampling exploits this smoothness. In particular, rather than uniformly drawing $N'$ samples from the whole search space $\mathcal{S}$, we sample locally around the best observed network architectures, using random mutation and crossover.

More specifically, the evolutionary sampling method is summarized in algorithm~\ref{alg:ev_sampling}. In each iteration $t=1,...,T$, we use $\mathcal{T}_t$ as
the population of observed network architectures, from which we pick $P$ parent architectures with the lowest true
validation error. We denote this parent set with $\mathcal{P}$. Until we reach the desired number of samples $N'$, we either perform
random mutation of one random architecture or a random crossover of two random architectures from
$\mathcal{P}$, while enforcing a given mutation/crossover ratio $\alpha \in [0,1]$. 

When we mutate an architecture $\mathcal{A}_1=(\mathbf{A}_1, \mathbf{X}_1)$, we randomly flip the binar value of each element of its adjacency matrix $\mathbf{A}_1$ and its vertice feature matrix $\mathbf{X}_1$ with the probability $p_{mutate} \in [0,1]$. For the crossover of two architectures $\mathcal{A}_1 = (\mathbf{A}_1, \mathbf{X}_1)$ and $\mathcal{A}_2=(\mathbf{A}_2, \mathbf{X}_2)$, we construct a new adjacency matrix, with elements that are either taken from $\mathbf{A}_1$ or from $\mathbf{A}_2$, and a new vertice feature matrix, with row vectors that are either taken from $\mathbf{X}_1$ or from $\mathbf{X}_2$, with probability $0.5$, respectively. 

\begin{algorithm}
    \caption{Evolutionary sampling of $\mathcal{S}'$.}
    \label{alg:ev_sampling}
     $\mathcal{P} \gets \mathrm{get\_best\_}P(\mathcal{T}_t, \mathcal{Y}_t)$\\
     $\mathcal{S}' \gets \{ \}$\\

    \While{$|\mathcal{S}'| < N'$}{
         $\mathcal{A}_1 \gets \mathrm{get\_rnd}(\mathcal{P})$\\
         $\mathcal{A}_2 \gets \mathrm{get\_rnd}(\mathcal{P})$\\
        \eIf{$|\mathcal{S}'| < \alpha N'$}{
             $\mathcal{S}' \gets \mathcal{S}' \cup \mathrm{mutate}(\mathcal{A}_1)$\\
        }{
             $\mathcal{S}' \gets \mathcal{S}' \cup \mathrm{crossover}(\mathcal{A}_1, \mathcal{A}_2)$\\
        }
    }
\end{algorithm}

\section{Sample efficiency gain for predictor based NAS}
\label{sec:gain}

In this section, we introduce the sample efficiency gain that we use to compare the different predictor-based NAS algorithms. 

To this end, consider that we sequentially pick, train and evaluate network architectures from the candidate set $\mathcal{C}$. At first, we compute the average number of network architectures that we need to train and evaluate this way, until we observe a validation error that is better or equal than a desired target $J_{target}$. Mathematically, this is an urn experiment, where we draw samples from $\mathcal{C}$ until success without replacement. Let 
\begin{align}
\mathcal{R}(J_{target}) =\{\mathcal{A}: \mathcal{A} \in \mathcal{C}, J_v(\mathcal{A}, \mathbf{w}^*) \leq J_{target}\}
\end{align}
be the subset of $\mathcal{C}$ that only contains architectures with $J_v(\mathcal{A}, \mathbf{w}^*) \leq J_{target}$. Further, let $K = |\mathcal{C}|$ and $M(J_{target}) = |\mathcal{R}(J_{target})|$.
We define the minimum number of trials $\tau(J_{target})$ until we observe $J_v(\mathcal{A}, \mathbf{w}^*) \leq J_{target}$ as
\begin{align}
	\tau(J_{target}) = \min \{k: A_k \in \mathcal{R}(J_{target}) \},
\end{align}
where $k$ is the trial index. From combinatorics we know that the probability that we draw an architecture with $J_v(\mathcal{A}, \mathbf{w}^*) \leq J_{target}$ from $\mathcal{C}$ after $k$ trials is
\begin{align}
	p_{\mathcal{C}}(\tau(J_{target})=k) &= \frac{(K-M(J_{target}))!}{(K-M(J_{target})-k+1)!} \nonumber \\
	                                    &\cdot \frac{(K-k)!}{K!}. \label{eq:p_tau_1}
\end{align}
Here, $\frac{(K-M(J_{target}))!}{(K-M(J_{target})-k+1)!}$ counts all possibilities 
how we can draw the first $k-1$ elements from $\mathcal{C} \setminus \mathcal{R}(J_{target})$ without success.
Further, there are $M(J_{target})$ possible ways that trial $k$ itself is a success. Everything is normalized by
$\frac{K!}{(K-k)!}$, i.e., by the total count of all possible ways in which we can choose $k$ elements from $\mathcal{C}$.

Equation~\ref{eq:p_tau_1} can be simplified to
\begin{align}
	p_{\mathcal{C}}(\tau(J_{target})=k) = \frac{M(J_{target})}{k} \frac{\binom{K-M(J_{target})}{k-1}}{\binom{K}{k}} \label{eq:p_tau_2}.
\end{align}
Further, using \ref{eq:p_tau_2}, we can now caluclate the desired average $\tau(J_{target})$, that is
\begin{align}
    E_{\mathcal{C}}[\tau(J_{target})] = \frac{K+1}{M(J_{target})+1}.
\end{align}

In the same way, we can calculate $E_{\mathcal{S}}[\tau(J_{target})]$ for the case that we do not draw architectures from $\mathcal{C}$, but from the original search space $\mathcal{S}$, instead. Because $\mathcal{S}$ is not narrowed down, using the predictor, it still contains many architectures with a high validation error. Therefore, we expect that $E_{\mathcal{S}}[\tau(J_{target})] >> E_{\mathcal{C}}[\tau(J_{target})]$.

Using these results, we define the sample efficiency gain $\mathrm{gain}(J_{target})$ in decibel as
\begin{align}
    \mathrm{gain}(J_{target}) = 10 \log \left ( \frac{E_{\mathcal{S}}[\tau(J_{target})]}{E_{\mathcal{C}}[\tau(J_{target})]} \right ).
\end{align}
In particular, $\mathrm{gain}(J_{target}$) measures how much faster we observe an architecture with
$J_v(\mathcal{A}, \mathbf{w}^*) \leq J_{target}$ if we pick from the candidate set $\mathcal{C}$, rather than from
the original search space $\mathcal{S}$. Therefore, it is directly related to the time that we save, by using a predictor to narrow down the search space.

Note, that the gain is $0$, if $E_{\mathcal{S}}[\tau(J_{target})]$ and $E_{\mathcal{C}}[\tau(J_{target})]$ are equal, meaning that we have to pick the same number of architectures from $\mathcal{S}$ and $\mathcal{C}$. This is meaningful, because in that case training networks from the candidate set does not give us any benefit. $\mathrm{gain}(J_{target})$ is easy to compute. We just have to estimate $M(J_{target})$ for $\mathcal{C}$ and $\mathcal{S}$ from samples.

\section{Experiments}
\label{sec:experiments}
We conduct experiments on the popular NASBench-101 dataset \cite{ying2019-bench101} that provides architecture/accuracy pairs for 423,624 unique network architectures. Each of these architectures is trained and evaluated three times on a separate test and validation set. They are defined by a layer adjacency matrix $\mathbf{A} \in \mathbb{R}^{7 \times 7}$ and a layer feature matrix $\mathbf{X} \in \mathbb{R}^{7 \times 5}$. NASBench-101 is designed to evaluate NAS algorithms in a controlled environment. As recommended in \cite{ying2019-bench101}, we use the reported validation accuracy in the search phase and evaluate the NAS algorithms based on the average test accuracy. In all our experiments, we compare to random search as a lower baseline, to the validation oracle that returns the architecture with the best validation accuracy and to the test oracle, that returns the architecture with the best test accuracy. 

For all experiments, our GCN predictor consists of 3 graph convolutional layers that extract 256 vertice features, i.e., $\mathbf{X}_l \in \mathbb{R}^{7 \times 256} \:\: l=1,...,3$, that are followed by a pooling and a fully-connected output layer. Further, we select $|\mathcal{T}_0| = 4$ random architectures from $\mathcal{S}$ to train the initial GCN predictor. In each iteration, we train the GCN for $2000$ epochs, using the pairwise cross-entropy loss given in sections~\ref{sec:pb_nas}. We use stochastic gradient descent with an initial learning rate $\lambda=0.01$ and a momentum $\gamma=0.9$. Further, the learning rate is decreased to zero, using a cosine-by-step schedule. After training, the trained predictor is used to narrow down the search space to $K=4$ candidate architectures, that are eventually trained, evaluated and added to $\mathcal{T}_t$ for the next iteration.

In our first experiment we investigate how the performance of the predictor-based algorithm degrades, if the predictor operates only on subsets 
of the full search space $\mathcal{S}$ to generate new architecture candidates $\mathcal{C}$. For this experiment, we use the most naive method to construct these subsets $\mathcal{S}'$, namely \emph{uniform sampling} as introduced in section~\ref{sec:pb_sampling}. 

\begin{figure}
    \centering
    \includegraphics[width=\linewidth]{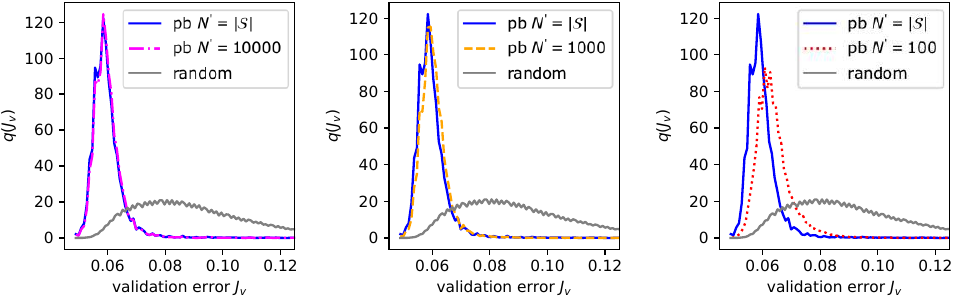}
    \caption{The distribution of the validation errors of the calculated architecture proposals, for the case that the predictor can use $N' = |\mathcal{S}|, 10000, 1000, 100$ elements from the search space to propose new candidates. Clearly, the distribution $q(J_v)$ is the narrowest and has the smallest mean if the predictor operates on the whole search space. Here, "pb" stands for predictor-based NAS.}
    \label{fig:hist_rnd}
\end{figure}

Figure~\ref{fig:hist_rnd}, shows the distribution of the true validation errors $J_v(\mathcal{A};\mathbf{w}^*)$ for network architectures from the proposed candidate set $\mathcal{C}$. In particular, we use different sizes of the subsets $\mathcal{S}'$. Namely $N' = |\mathcal{S}|, 10000, 1000, 100$. As a reference, we also show the case of random search (grey). In this case, no predictor is used at all, meaning that $\mathcal{C}$ is drawn randomly from $\mathcal{S}'$.

For the random case, the distribution of the validation errors $q(J_v)$ is very broad and has long tails. Hence, it is very unlikely that $\mathcal{C}$ contains a network architecture with a low $J_v$. If the predictor operates on the full search space, 
i.e. for $N' = |\mathcal{S}|$ drawn in blue, $q(J_v)$ the narrowest and has the smallest mean. This means, that the predictor correctly discards architectures with a large $J_v$. However, for decreasing $N'$, $q(J_v)$ again gets broadened and its mean gets shifted towards larger $J_v$.

\begin{figure}
    \centering
    \includegraphics[width=\linewidth]{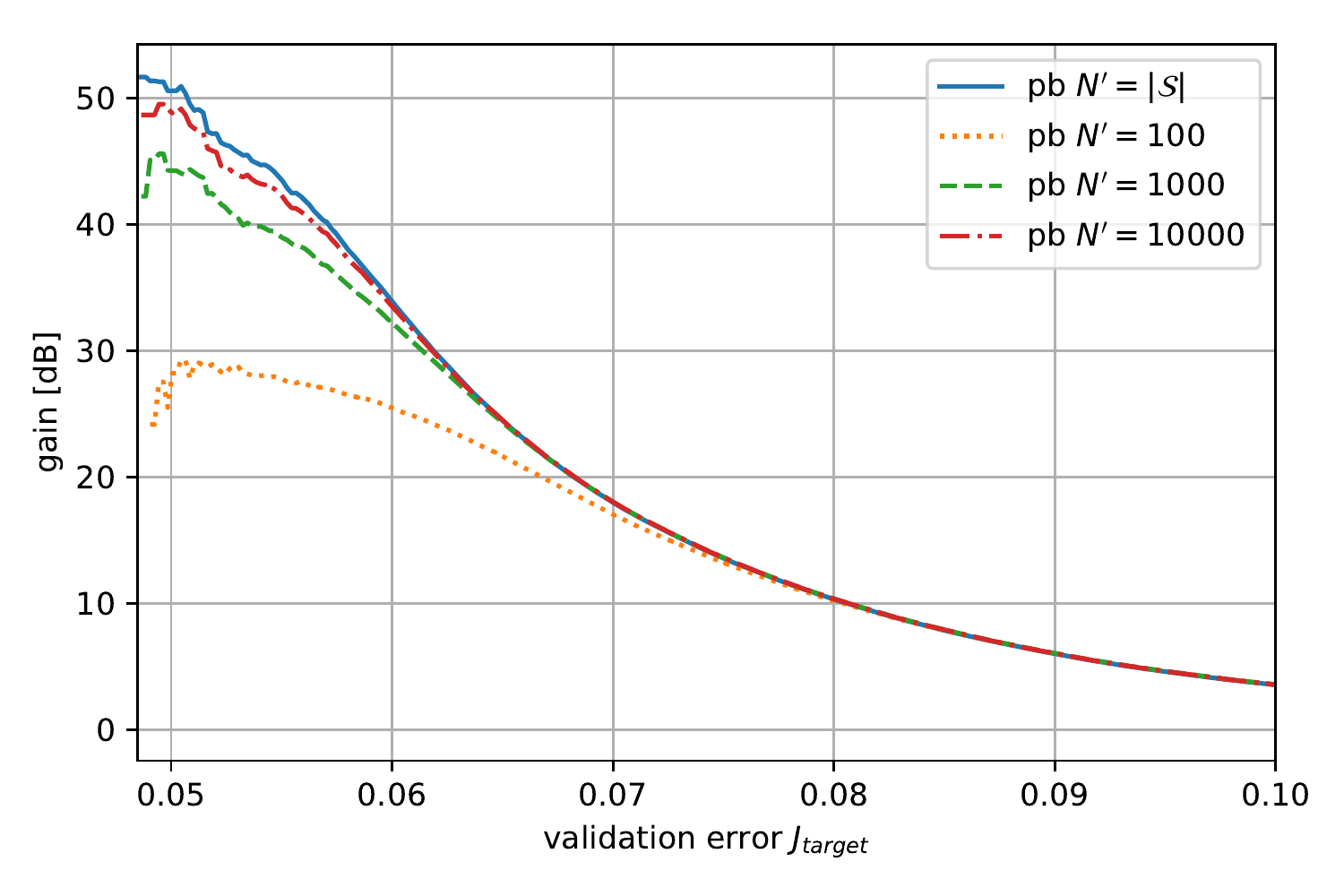}
    \caption{The sampling gain, as defined in section~\ref{sec:gain}, for the case that the predictor can use $N' = |\mathcal{S}|, 10000, 1000, 100$ elements from the search space to propose new candidates.}
    \label{fig:gain}
\end{figure}

Figure~\ref{fig:gain} shows the corresponding gain that we defined in section~\ref{sec:gain}. The highest gain is clearly
reached for $N' = |\mathcal{S}|$, i.e., if the predictor uses the whole search space. Interestingly, the gain is largest for small $J_{target}<0.05$.
Here, the predictor-based algorithm reaches a gain of about $50dB$, meaning that it is roughly $100000$ times more sample efficient than random search. As expected, the gain decreases proportional to $N'$. Note, that the reduction is not the same for all $J_{target}$. In fact, the gain is reduced most for $0 \leq J_{target} \leq 0.05$, while
remaining nearly unchanged for $J_{target}>0.07$. $\mathrm{gain}(0.05)$ is for example $25dB$ smaller if we use $N'=100$ instead of $N'=\mathcal{S}$. This means, that the sample efficiency of the predictor-based algorithm is roughly reduced by a factor $300$ at that point. 
Obviously, the choice of $N'$ has a big influence on the performance of predictor-based algorithms. In particular for a small $N'$, the predictor looses its ability to propose network architectures with a very low validation error. The effects of a small $N'$ therefore can be observed best late during optimization, when we approach the optimal network architectures.

\begin{figure}
    \centering
    \includegraphics[width=\linewidth]{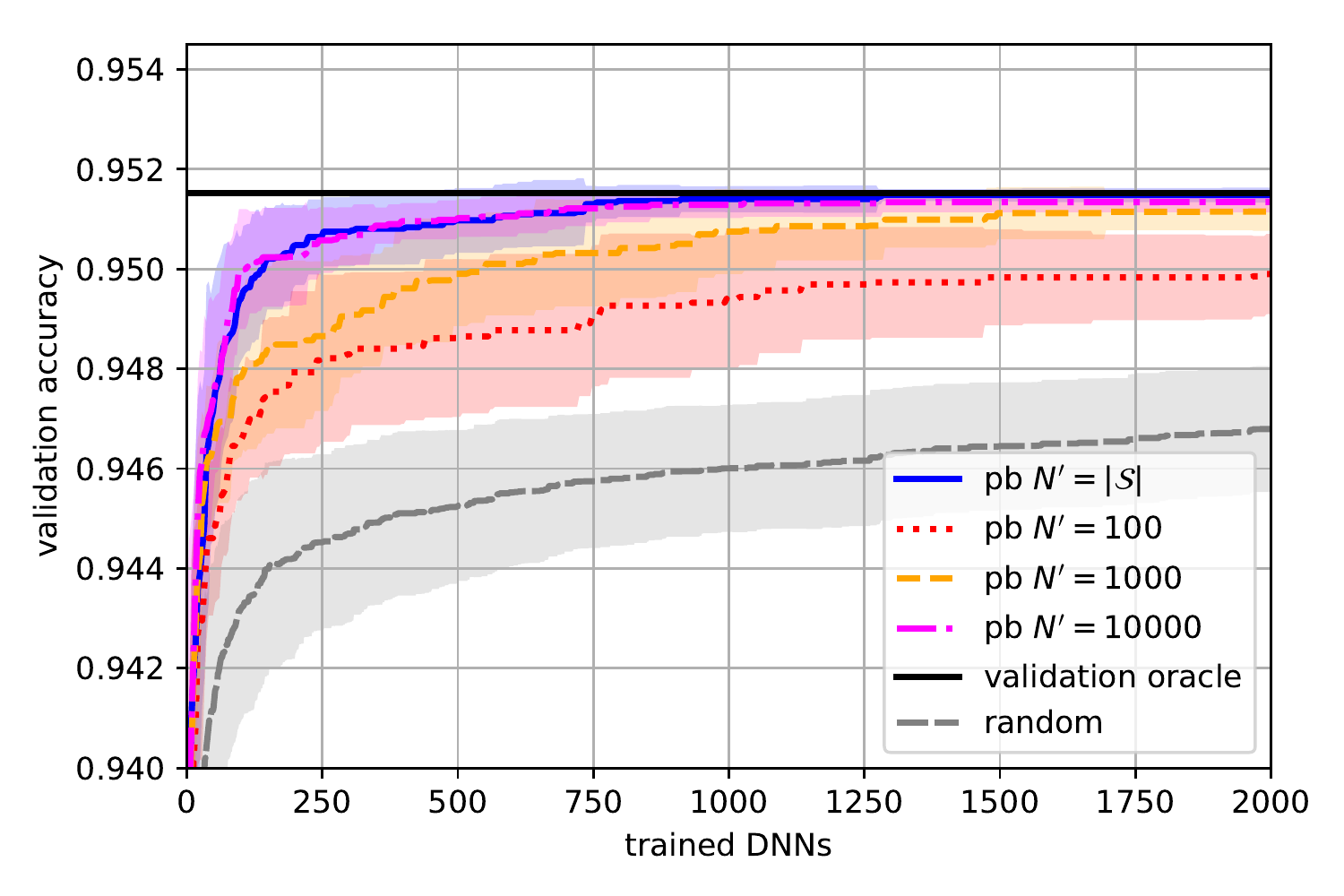}
    \caption{Convergence of the validation error, for the case that the predictor can use $N' = |\mathcal{S}|, 10000, 1000, 100$ elements from the search space to propose new candidates.}
    \label{fig:val_convergence_pb}
\end{figure}

\begin{figure}
    \centering
    \includegraphics[width=\linewidth]{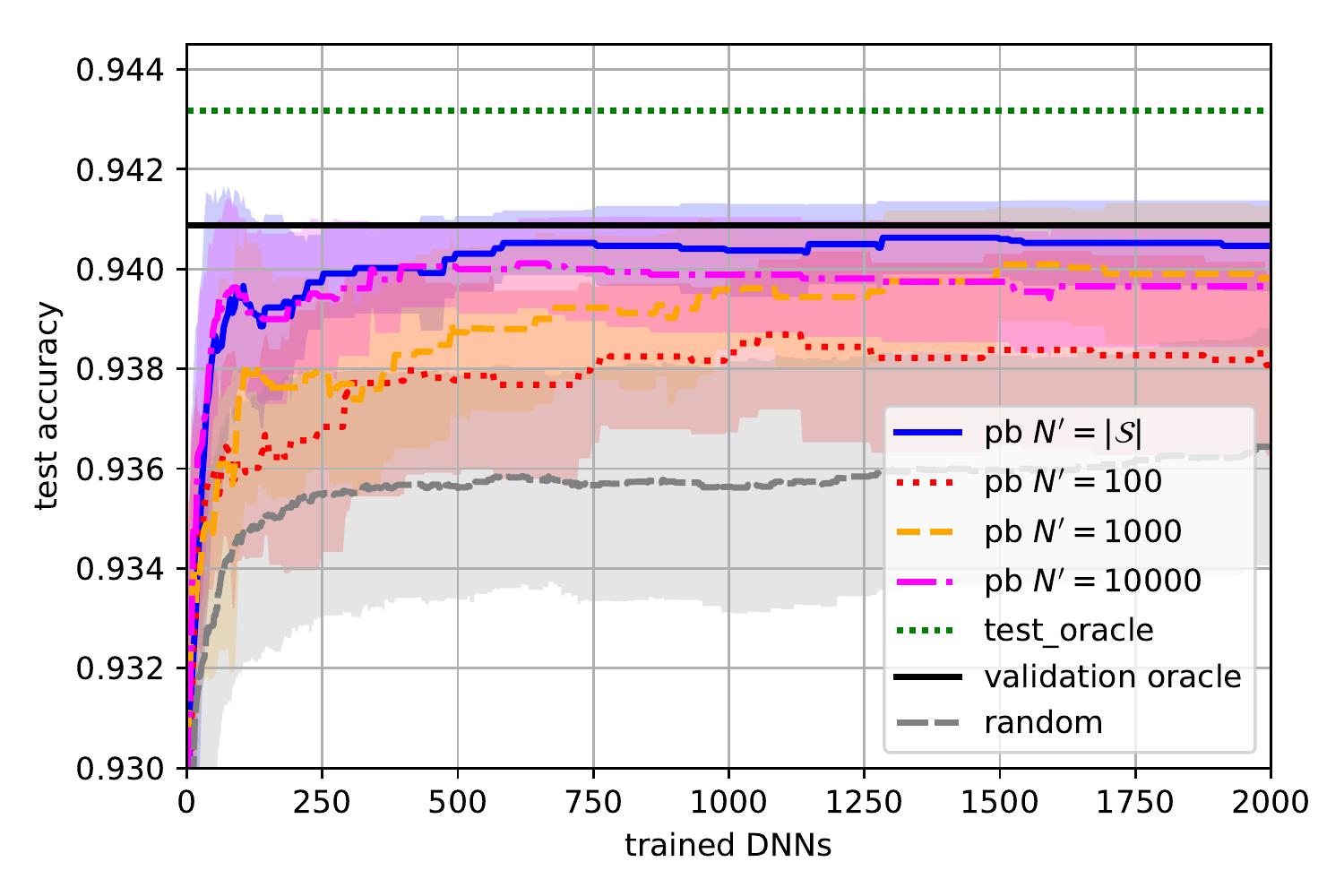}
    \caption{Convergence of the test error, for the case that the predictor can use $N' = |\mathcal{S}|, 10000, 1000, 100$ elements from the search space to propose new candidates.}
    \label{fig:test_convergence_pb}
\end{figure}

Finally, figures~\ref{fig:val_convergence_pb} and \ref{fig:test_convergence_pb} show how this reduced sampling gain affects
the average convergence of the validation and the test error. For this plot, we averaged over $20$ different runs and display the mean and the standard deviation. In addition to the random baseline, we also give the validation and the test oracle, which always return the best validation and test accuracy that exists in the whole dataset. Both act as an upper bound for the performance of the NAS algorithm. While we quickly approach the validation oracle for $N' = |\mathcal{S}|$ and $10000$, the convergence of both the validation and test error is significantly worse for $N'= 1000$ and $100$. In particular, we cannot approach the validation oracle for $N'= 1000$ and $100$.

\begin{figure}
    \centering
    \includegraphics[width=\linewidth]{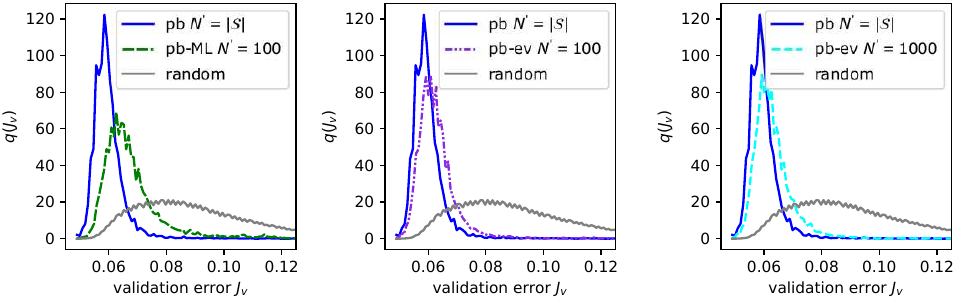}
    \caption{The distribution of the validation errors of the proposed candidate networks, for the case that the predictor can use $N' = 1000, 100$ elements from the search space to propose new candidates.}
    \label{fig:hist_sample}
\end{figure}

In our second experiment we want to answer the question if we can increase the sample efficiency of the predictor-based algorithms that operate on small subsets $\mathcal{S}'$ of the search space. Therefore, we change the way how the subsets $\mathcal{S}'$ are constructed. In particular, we use $N'=100$ or $1000$ and construct $\mathcal{S}'$ using the ML-based and the evolutionary based sampling, that we introduced in section~\ref{sec:pb_sampling}. Figure~\ref{fig:hist_sample}, shows the distribution of the validation errors in $\mathcal{C}$. For both sampling methods. Here, "pb-ML" and "pb-ev" denote predictor-based algorithms that use ML-based and evolutionary based sampling, respectively. Both methods result in distributions $q(J_v)$ that are considerably narrower than the random case. The best result can be obtained, using evolutionary sampling. Here, the distribution is almost as narrow as for $N'=|\mathcal{S}|$.

\begin{figure}
    \centering
    \includegraphics[width=\linewidth]{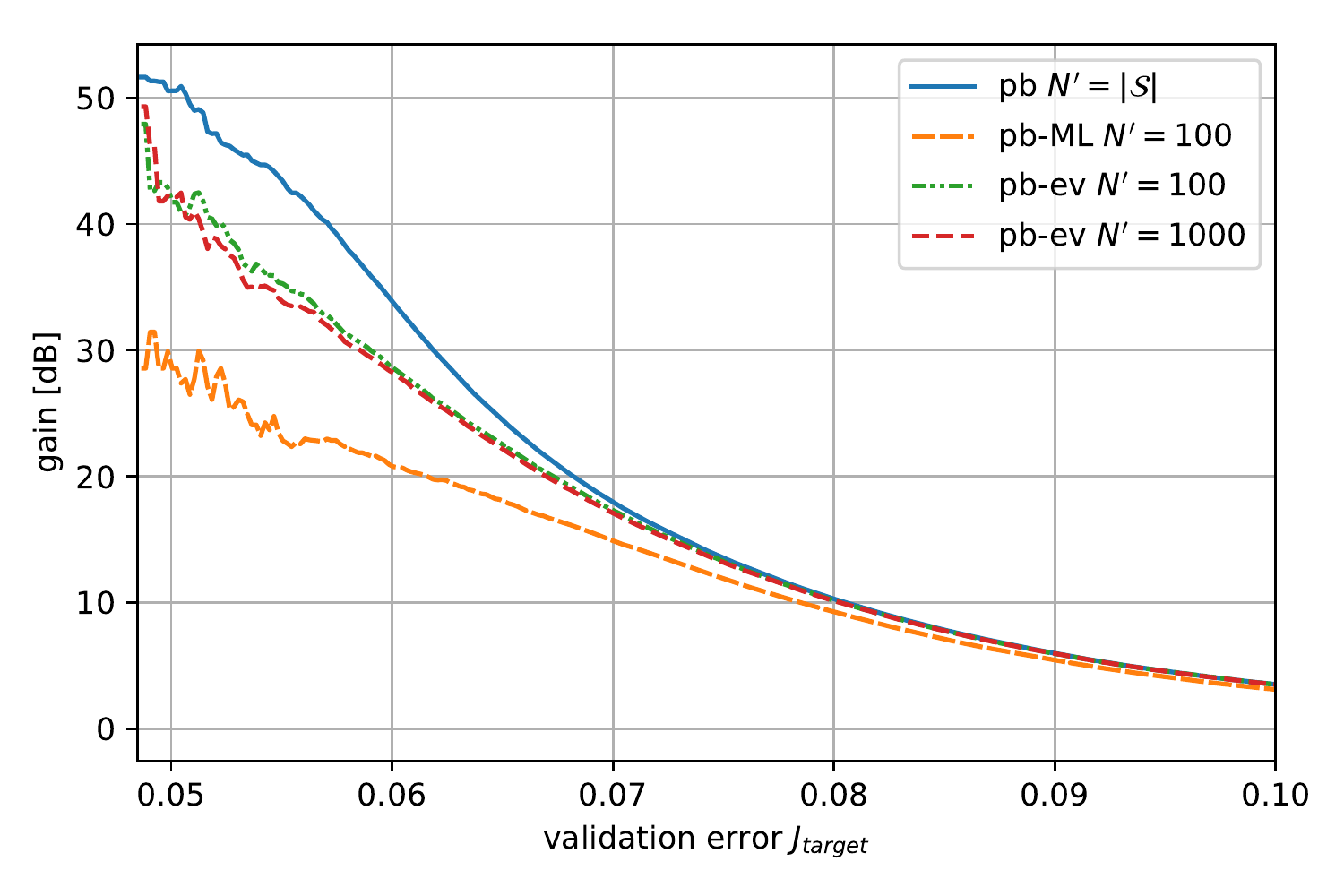}
    \caption{The sampling gain, as defined in section~\ref{sec:gain}, for the case that the predictor can use $N' = 1000, 100$ elements from the search space to propose new candidates.}
    \label{fig:gain_sample}
\end{figure}

Figure~\ref{fig:gain_sample} shows the related sample efficiency gain. In comparison to the uniform case from figure~\ref{fig:gain}, pb-ML performs a bit worse than uniform sampling for $J_{target}>0.05$. However, with about $30dB$ gain, ML-based sampling achieves a slightly higher gain for $J_{target}<0.05$ when we compare it to the uniform case that only achieves about $25dB$ gain.
Most promising is evolutionary sampling. For $J_{target}<0.05$, it approaches a gain of $50dB$ and therefore is comparable to the performance of predictor-based with $N'=|\mathcal{S}|$.

\begin{figure}
    \centering
    \includegraphics[width=\linewidth]{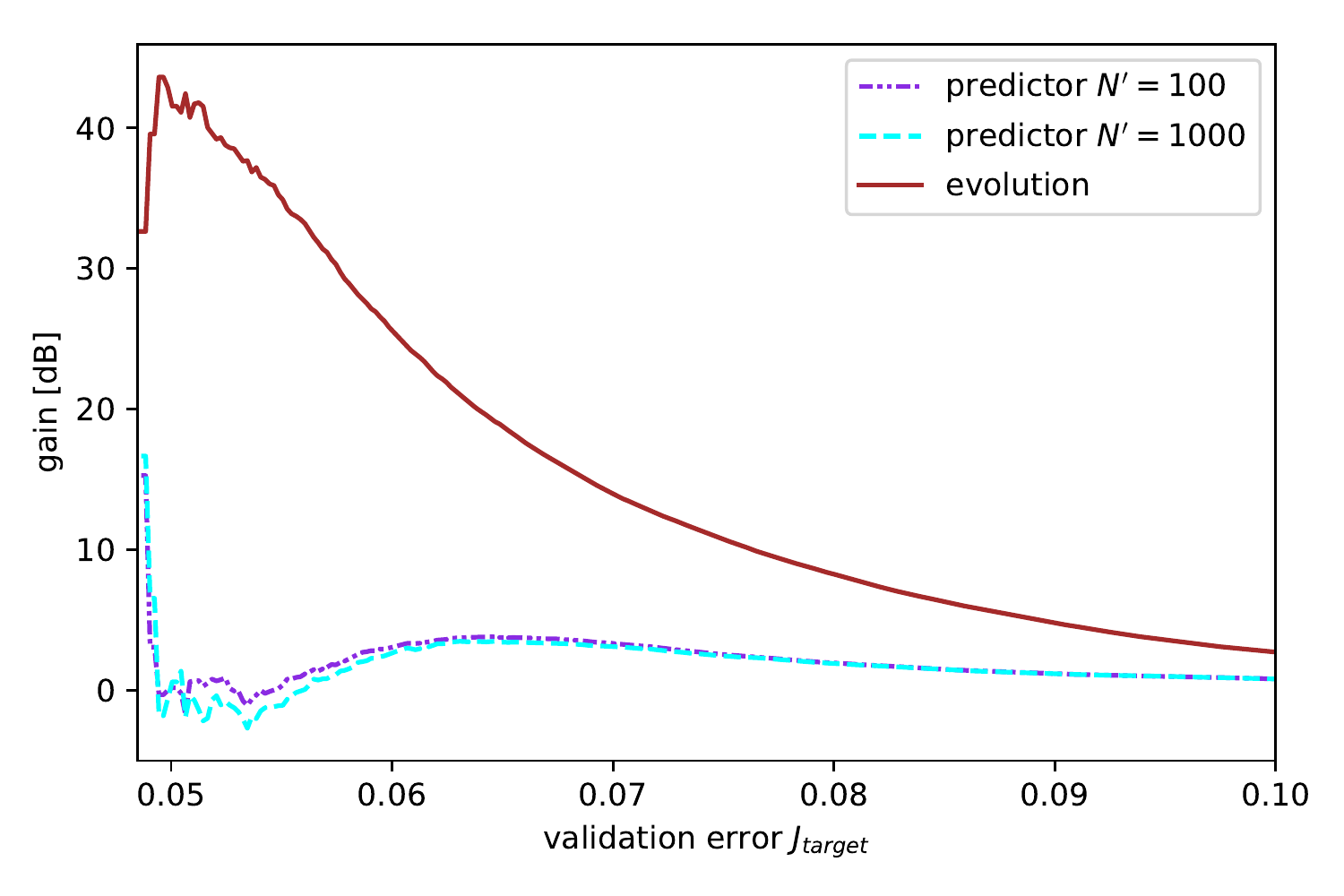}
    \caption{Separating the gain caused by the predictor and the gain caused by evolutionary sampling.}
    \label{fig:gain_splitted}
\end{figure}

In figure~\ref{fig:gain_splitted}, we further split the gain into the gain that is due to evolutionary sampling and the gain that is caused 
by the predictor. More specifically, following section~\ref{sec:gain}, we plot both 
\begin{align}
    \mathrm{gain}_e(J_{target}) &= 10 \log \left ( \frac{E_{\mathcal{S}}[\tau(J_{target})]}{E_{\mathcal{S}'}[\tau(J_{target})]} \right )\\
    \mathrm{gain}_p(J_{target}) &= 10 \log \left ( \frac{E_{\mathcal{S}'}[\tau(J_{target})]}{E_{\mathcal{C}}[\tau(J_{target})]} \right )
\end{align}. 
We can see that evolutionary sampling alone already results in a high $\mathrm{gain}_e(J_{target})$ for $J_{target}>0.05$. However, we also observe that $\mathrm{gain}_e(J_{target})$ decreases for small $J_{target}<0.05$.
Luckily, $\mathrm{gain}_p(J_{target})$ seems to have an reciprocal behaviour. 
In particular, $\mathrm{gain}_p(J_{target})$ is especially large for $J_{target}<0.05$. Therefore, the proposed combination of evolutionary sampling with predictor-based NAS seems to be a perfect match.

\begin{figure}
    \centering
    \includegraphics[width=\linewidth]{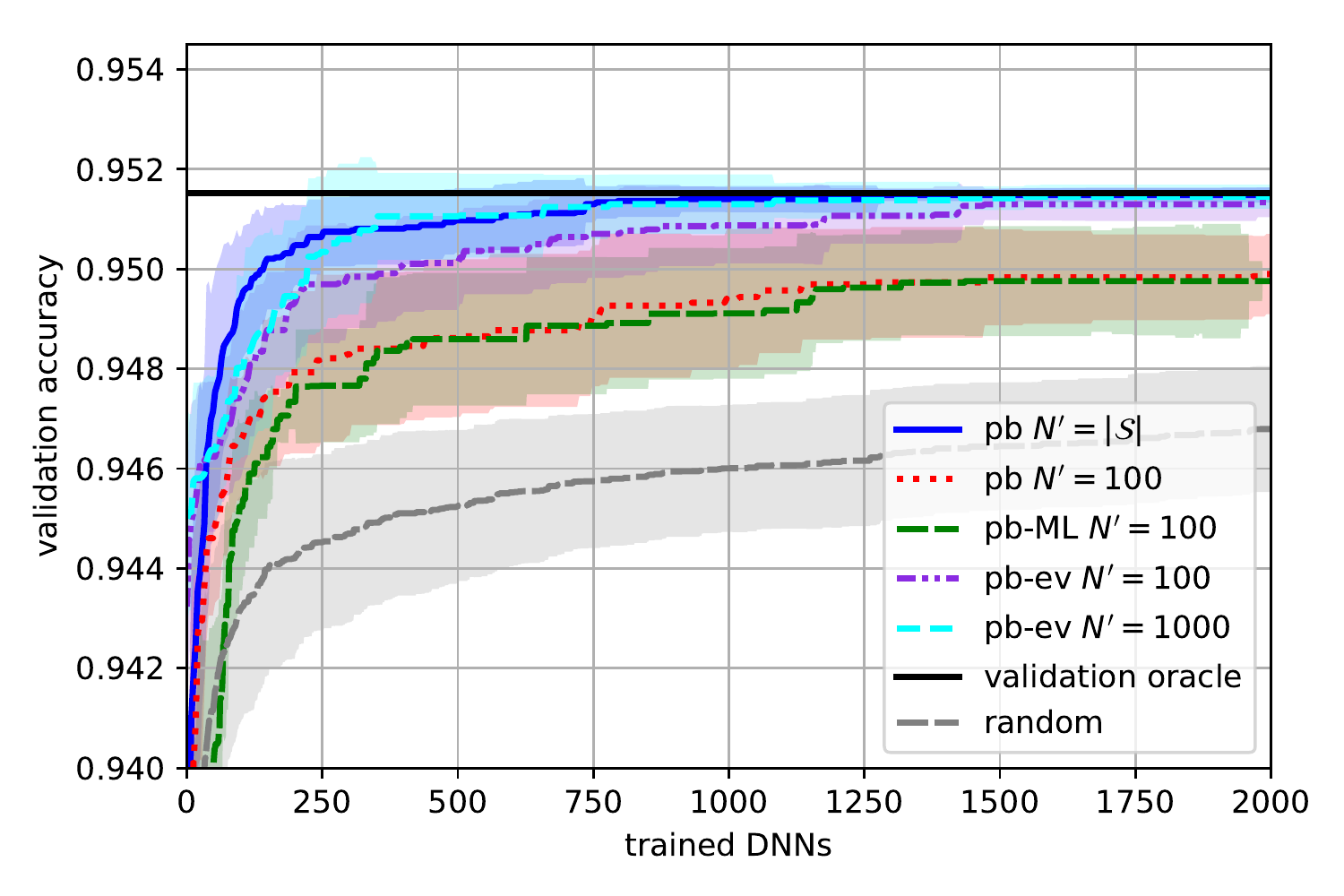}
    \caption{The convergence of the validation error for predictor-based NAS, using ML-based or evolutionary sampling.}
    \label{fig:convergence_pb_sampling}
\end{figure}

\begin{figure}
    \centering
    \includegraphics[width=\linewidth]{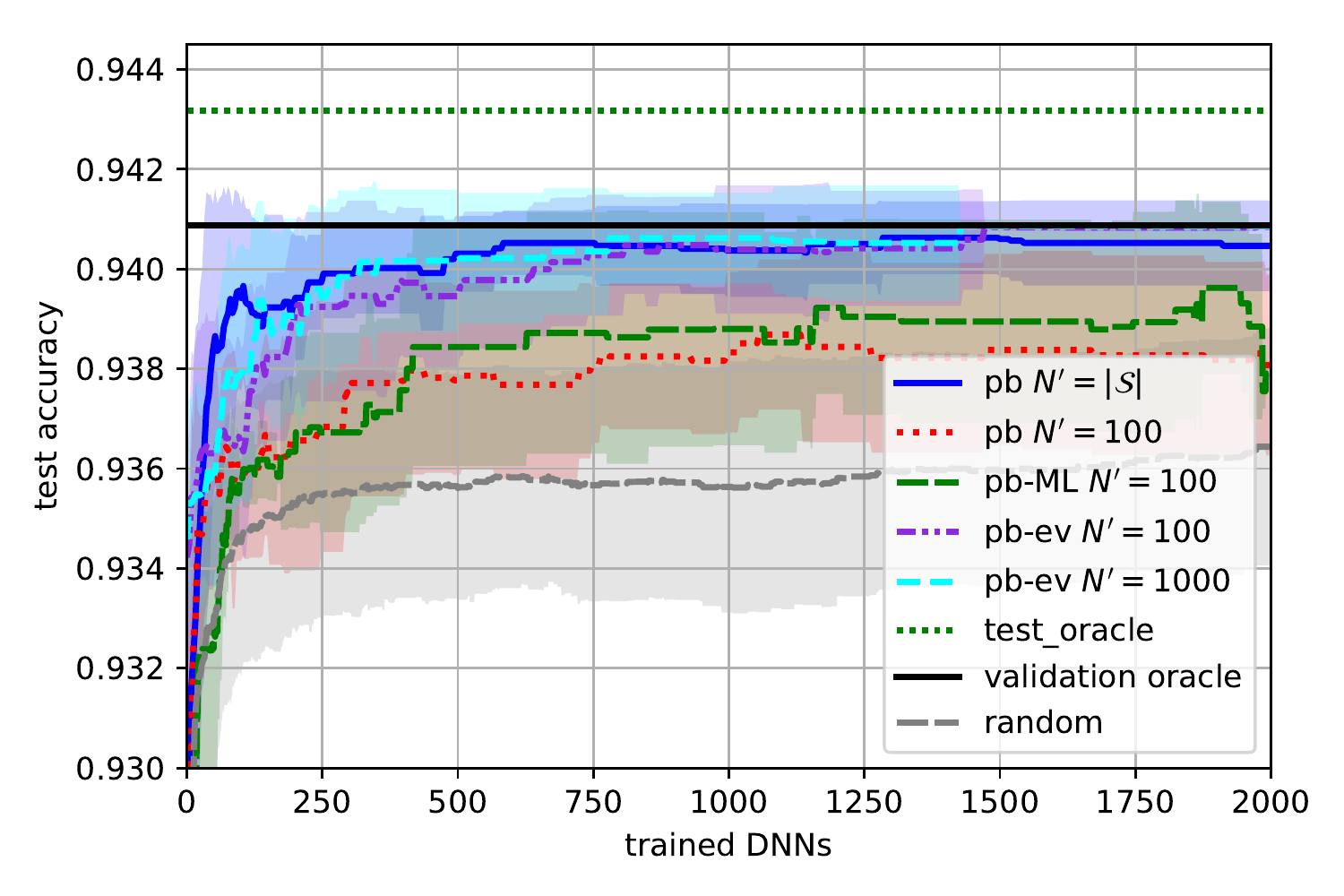}
    \caption{The convergence of the test error for predictor-based NAS, using ML-based or evolutionary sampling.}
    \label{fig:test_convergence_pb_sampling}
\end{figure}

Finally, figures~\ref{fig:convergence_pb_sampling} and \ref{fig:test_convergence_pb_sampling} show the the corresponding average convergence of 
the validation and the test error, respectively. While pb-ML does not converge any better than if we use predictors that operate on uniformly sampled subsets of $\mathcal{S}$, 
using pb-ev obviously performs very well. Infact, pb-ev converges almost identical to the baseline algorithm that uses $N'=|\mathcal{S}|$. All runs converged to the validation oracle after having observed about $1400$ network architectures. That is comparable to the results 
from \cite{wen2019neural}. However, pb-ev requires considerably less computational effort. We conclude, that combining evolutionary sampling with predictor-based search results in architecture search algorithms that are sample efficient and that scales well to large search spaces.

\section{Conclusion}
In this paper, we have introduced the sample efficiency gain, that can be used to evaluate and compare different
predictor-based NAS algorithms. We demonstrated that the performance of predictor-based NAS algorithms is severely reduced if the predictor cannot use the whole search space $\mathcal{S}$ to propose new candidate architectures and needs to operate on subsets of $\mathcal{S}$, instead. In particular, our experiments on NASBench-101 show, that the sample efficiency of a predictor-based algorithm can be reduced by a factor of $300$, if the predictor uses only very small subset of the search space. In practice, that is a big problem because NAS algorithms often must operate on very large search spaces that contain billions of candidate architectures. To address this problem, we demonstrated that we can carefully design random subsets of the search space using evolutionary sampling. In particular, predictor-based algorithm that operate on these random subsets have a similar performance than those that operate on the whole search space. This is a very important step to make predictor-based algorithms suited for real-world NAS problems.

{\small
\bibliographystyle{ieee_fullname}
\bibliography{egbib}
}

\end{document}